\documentclass[conference]{IEEEtran}
\ifCLASSINFOpdf
\else
\fi
\usepackage{amsmath}
\usepackage{amsfonts}
\usepackage{mathtools}
\usepackage{bm}

\usepackage{url}

\usepackage{xcolor, colortbl}
\usepackage{hhline}
\usepackage{graphicx}
\usepackage{mdframed}
\usepackage{bbding}
\usepackage{color, soul}
\usepackage{multirow}
\usepackage{soul}  

\usepackage{tikz}
\usetikzlibrary{shapes.geometric,calc}

\newcommand\numberthis{\addtocounter{equation}{1}\tag{\theequation}}

\newcommand\score[2]{
\pgfmathsetmacro\pgfxa{#1+1}
\tikzstyle{scorestars}=[star, star points=5, star point ratio=2.25, draw,inner sep=0.15em,anchor=outer point 3]
\begin{tikzpicture}[baseline]
  \foreach \i in {1,...,#2} {
    \pgfmathparse{(\i<=#1?"black":"white")}
    \edef\starcolor{\pgfmathresult}
    \draw (\i*1em,0) node[name=star\i,scorestars,fill=\starcolor]  {};
   }
   \pgfmathparse{(#1>int(#1)?int(#1+1):0}
   \let\partstar=\pgfmathresult
   \ifnum\partstar>0
     \pgfmathsetmacro\starpart{#1-(int(#1))}
     \path [clip] ($(star\partstar.outer point 3)!(star\partstar.outer point 2)!(star\partstar.outer point 4)$) rectangle 
    ($(star\partstar.outer point 2 |- star\partstar.outer point 1)!\starpart!(star\partstar.outer point 1 -| star\partstar.outer point 5)$);
     \fill (\partstar*1em,0) node[scorestars,fill=black]  {};
   \fi

,\end{tikzpicture}
}

\makeatletter
\newcommand{\thickhline}{%
    \noalign {\ifnum 0=`}\fi \hrule height 1pt
    \futurelet \reserved@a \@xhline
}
\makeatother

\definecolor{negativeRed}{RGB}{192, 0, 0}
\definecolor{positiveGreen}{RGB}{0, 120, 40}
\definecolor{lg}{RGB}{200, 200, 200}

\newcommand{\R}{\mathbb{R}}

\begin{document}
%
\title{Multi-Aspect~Sentiment~Analysis with Latent~Sentiment-Aspect~Attribution}

\author{\IEEEauthorblockN{Yifan Zhang}
\IEEEauthorblockA{
University of Houston\\
aeryen@gmail.com}
\and
\IEEEauthorblockN{Fan Yang}
\IEEEauthorblockA{
University of Houston\\
fyang11@uh.edu}
\and
\IEEEauthorblockN{Marjan Hosseinia}
\IEEEauthorblockA{
University of Houston\\
mhosseinia@uh.edu}
\and
\IEEEauthorblockN{Arjun Mukherjee}
\IEEEauthorblockA{
University of Houston\\
arjun@uh.edu}

}


%






\maketitle

\begin{abstract}
In this paper, we introduce a new framework called the sentiment-aspect attribution module (SAAM). SAAM works on top of traditional neural networks and is designed to address the problem of multi-aspect sentiment classification and sentiment regression. The framework works by exploiting the correlations between sentence-level embedding features and variations of document-level aspect rating scores. We demonstrate several variations of our framework on top of CNN and RNN based models. Experiments on a hotel review dataset and a beer review dataset have shown SAAM can improve sentiment analysis performance over corresponding base models. Moreover, because of the way our framework intuitively combines sentence-level scores into document-level scores, it is able to provide a deeper insight into data (e.g., semi-supervised sentence aspect labeling). Hence, we end the paper with a detailed analysis that shows the potential of our models for other applications such as sentiment snippet extraction.
\end{abstract}


%
\IEEEpeerreviewmaketitle

\section{Introduction}
\label{Introduction}
 
In this work, we propose a novel neural-network-based framework called a sentiment-aspect attribution module (SAAM) to solve the problem of document level multi-aspect sentiment analysis (MASA).

The proposed SAAM module can be trained using a set of documents tagged with overall and aspect ratings. During inference, SAAM employs \emph{latent sentiment-aspect attribution} (LSAA) mechanism, where it assigns a latent aspect distribution to each sentence while also estimates their sentiment scores. The estimated latent aspect distribution and sentiment scores for each sentence of a document are then pooled together to estimate the document-level review ratings.

To our knowledge, the proposed idea is the first neural network model capable of discovering both sentiment and aspect information at the sentence level with only document-level aspect-rating labels. Moreover, the framework we introduced in this work is not a single, specific neural network architecture. Instead, it is an add-on component that can be added to other popular neural network architectures to support MASA and LSAA. 

\begin{figure}[t]
\begin{mdframed}
\footnotesize
$\bullet$ {\color{negativeRed} Definitely not a 5 star resort I’m dumbfounded that this hotel gets good reviews and is so highly rated. \texttt{[1.23, Value]}}
$\bullet$ {\color{negativeRed} It's decidedly a 3 star property, not 5 stars as indicated. \texttt{[-0.04, Service]}}
$\bullet$ {\color{negativeRed} The rooms are very dated and run down, old crappy beds and pillows, an old tv and overall poorly maintained. \texttt{[-2.97, Room]}}
$\bullet$ {\color{negativeRed} The whole property is pretty run down and old–looking. \texttt{[-0.47, Location]}}
$\bullet$ {\color{negativeRed} The food is subpar, not one meal I had would be called great. \texttt{[-2.23, Service]}}
$\bullet$ {\color{negativeRed} The service is uneven and the staff is poorly trained and uninformed. \texttt{[-2.23, Service]}}
$\bullet$ {\color{positiveGreen} The beach is great, it's the only redeeming factor. \texttt{[1.27, Location]}}
$\bullet$ {\color{negativeRed} However the resort is a 1-hour taxi trip from the airport. \texttt{[1.68, Location]}}

\vspace{0.2cm}
\begin{tabular}{llll}
\textbf{Overall:} & \score{2}{5} & Value: & \score{1}{5} \\
Room: & \score{1}{5} & Location: & \score{4}{5} \\
Cleanliness: & \score{2}{5} & Service: & \score{2}{5} \\
\end{tabular}

\end{mdframed}
\caption{A sample hotel review with user submitted ratings shown beneath. Sentiment scores and aspects assigned to sentences by our model in brackets.}
\label{hotel sample}
\end{figure}

\begin{figure}[t]
\begin{mdframed}
\footnotesize
$\bullet$ {\color{negativeRed} This beer is yellow, fizzy, and clearly meant for washing dirt out of your mouth after mowing the lawn. \texttt{[1.035, Appearance]}}
$\bullet$ {\color{negativeRed} I'm not even sure it's good for that. \texttt{[3.245, Taste]}}
$\bullet$ {\color{negativeRed} It's definitely yellow and fizzy, with no head to speak of, and zero lacing. \texttt{[-1.27, Appearance]}}
$\bullet$ {\color{negativeRed} It almost smells like a loaf of bread, and nearly tastes the same. \texttt{[4.255, Aroma]}}
$\bullet$ {\color{negativeRed} It's very earthy and grainy with nary a hop to be found. \texttt{[3.58, Taste]}}
$\bullet$ {\color{negativeRed} Man, I love me some Caldera, but I would rather drink a Bud Light than this on a hot summer day. \texttt{[1.845, Appearance]}}
$\bullet$ {\color{negativeRed} Sorry guys, but this beer gets an F. \texttt{[1.495, Taste]}}

\vspace{0.2cm}
\begin{tabular}{llll}
\textbf{Overall:} & \score{1}{5} & & \\
Appearance: & \score{1}{5} & Taste: & \score{1}{5} \\
Palate: & \score{1}{5} & Aroma: & \score{1.5}{5} \\
\end{tabular}

\end{mdframed}
\caption{A sample beer review with user submitted ratings shown beneath. Sentiment scores and aspects assigned to sentences by our model in brackets.}
\label{beer sample}
\end{figure}

Fig.\ref{hotel sample} and Fig.\ref{beer sample} show two sample reviews from the hotel review and beer review datasets. The latent aspect and sentiment score of each sentence estimated using our SAAM framework are displayed in brackets. We build 4 variations of our SAAM framework (two classification and two regression) to demonstrate the possibilities available. We stack these 3 variations of SAAM on top of a CNN \cite{kim2014convolutional} and a GRU-based RNN to demonstrate the framework's ability to generalize, as well as compare the performance between them.

Experimental results on the TripAdvisor hotel review dataset and BeerAdvocate beer review dataset show the effectiveness of the proposed approaches by showing performance improvement over corresponding base models as well as other baselines on several metrics for classification and regression variations of the MASA task. Additionally, we evaluate our model's ability of attributing aspect label to each sentence of a document by using manually labeled data as well as a heuristic keyword approach. We publish these processed datasets as well as sentence aspect labeling to better promote researches in this novel task.

In the last section of this paper, we explore a novel capability of the proposed model --- extracting sentiment snippets for each aspect along with a score rating. This is a useful by-product of the model and can be helpful in other tasks such as summarization.

\section{Backgrounds and Motivations}
\label{sec:bg}

\subsection{Related Works}

In \cite{wang2010latent}, the multi-aspect rating task was performed using generative modeling. Later, in \cite{wang2011latent}, a unified generative model for rating analysis was proposed that did not require explicit aspect keyword supervision. However, the model does not utilize aspect ratings of a document but instead uses overall ratings to discover latent aspects and estimate ratings on each aspect. A supervised LDA-like \cite{mcauliffe2008supervised} scheme was proposed in \cite{titov2008joint} and later in \cite{lu2011multi} that regressed the local and global topics (aspects) of reviews with the overall and aspect ratings for each review. Ranking algorithms are designed to either identify important aspects \cite{yu2011aspect} or aspect rating prediction without discovering them \cite{snyder2007multiple}. Another model used document-level multi-aspect ratings as a form of ``weak supervision'' to uncover sentence aspects. While it was quite successful in the sentence aspect attribution task, its primary purpose was not intended for estimating sentiment rating \cite{mcauley2012learning}.

There are many research efforts around SemEval 2015 and 2016 ABSA dataset \cite{pontiki2016semeval}. In these datasets, both aspect and sentiment polarity labels are available at both sentence and document levels. As such, the works such as \cite{tang2016effective,ruder2016hierarchical} address a somewhat different problem than the one in this paper. Both of these works utilize the sentence level labels that are not truly available in real-world review datasets and require labor-intensive labeling. Furthermore, the datasets include an extensive set of aspect categories. In contrast, real-world datasets such as the TripAdvisor review dataset have a fixed small set of aspects (roughly 3 - 5) that users rate on at the document level. These small differences in problem setting ultimately lead to very different solutions and models, and we believe both problem settings have their values.

In recent years, deep learning-based models have dramatically changed the field of natural language processing and significantly improved the performance of document classification \cite{yang-etal-2016-leveraging, boumber2018experiments, zhang2020birds}, Machine Translation \cite{bahdanau2016neural}, and Language Modeling \cite{devlin2019bert}. Convolutional neural networks (CNN) \cite{kim2014convolutional}, recurrent neural networks such as LSTM/GRU \cite{tang2016effective, ruder2016hierarchical} and more recently, pre-trained Bert based architecture have been proposed to solve the problem of sentiment analysis, and these models have significantly advanced state-of-the-art. Pre-trained transformer-alike architectures such as BERT with an extra task-specific layer are fine-tuned on domain reviews for aspect extraction and sentiment classification separately \cite{xu-etal-2019-bert}.

\subsection{Why SAAM}
Most of these previously mentioned deep learning classification models are built on top of some form of \textit{base models} (also called \textit{encoders}) that can take in an embedded sequence of text, and connect them to one or several layers of fully connected layers (also called \textit{decoders} or \textit{classification heads}) to ultimately estimate the probability distribution at the document level. While much progress is made in improving these base models' expressiveness, little attention was paid to the connection between token level outputs of the base models and the final prediction outputs. These token-to-doc connections are often either done by max-pooling/average-pooling \cite{kim2014convolutional,jeremy2018} or directly use the last/first token's output embedding as the document level embedding \cite{ruder2016hierarchical, xu-etal-2019-bert}.

We believe these existing token-to-doc connection schemes are not expressive enough and can become an information bottleneck in both the training and inferencing stage. In comparison, our SAAM framework provides an expressive connection between each sentence and the document level outputs. In doing so, the SAAM framework can further estimate the latent aspect distribution in each sentence, along with their sentiment rating score. Such fine-grained analysis capability, which we refer to as LSAA, provides more insight into the data because typical document-level sentiment classification or regression is a unison of sentiments expressed in various sentences across different aspects.

Secondly, our model only requires overall and aspect document-level ratings during the training stage, which can be acquired by most online review systems that use formats similar to those illustrated in Fig.\ref{hotel sample} and \ref{beer sample}. This means that architecture does not require any sentence-level aspect or sentiment supervision and can be easily applied to most existing review datasets and systems.

Lastly, by assigning each sentence to a proper aspect, the SAAM framework's LSAA capability will allow the generation of aspect-specific sentiment snippets. This feature is similar to a summarization system, where the summarization is based on choosing the relevant sentence under different latent aspects. These three major differences not only allow our model to improve upon current MASA methods, but also take into account variations of sentiment analysis tasks under different perspectives.

\section{Sentiment-Aspect Attribution Module}
\label{models}

\subsection{Problem Formulation}

Formally, we will refer to the text content part of a review simply as \textit{review} in the remaining part of this paper, and denote a single review using $r$. We use $s_i$ to refer to the $i$th sentence of a document and document is thus consisting of $\left|s\right|$ number of sentences. The set of factors that can be evaluated and rated by a reviewer are referred to as aspects, denoted using $A$. And $\lvert A \rvert$ is used to denote the cardinality of set $A$.
For example, the hotel review data we are working with:
$$A=\left\{Value,Room,Location,Cleanliness,Service\right\}$$

The actual overall rating and aspect ratings associated with a review $r$ are denoted as $R_{overall} (r)$ and $R_{aspects} (r)$. To correspond to the 5-star rating scheme, we assume that overall rating is scalar and aspect ratings is a vector consisting of $\lvert A \rvert$ number of elements:
$R_{overall} (r) \in \left\{1,2,3,4,5\right\}$ and $R_{aspects} (r) \in \left\{1,2,3,4,5\right\}^{\left|A\right|}$


\subsection{SAAM Classification--1 (SAAM-C1)}
\label{C1}

\begin{figure}[t]
    \includegraphics[width=\linewidth]{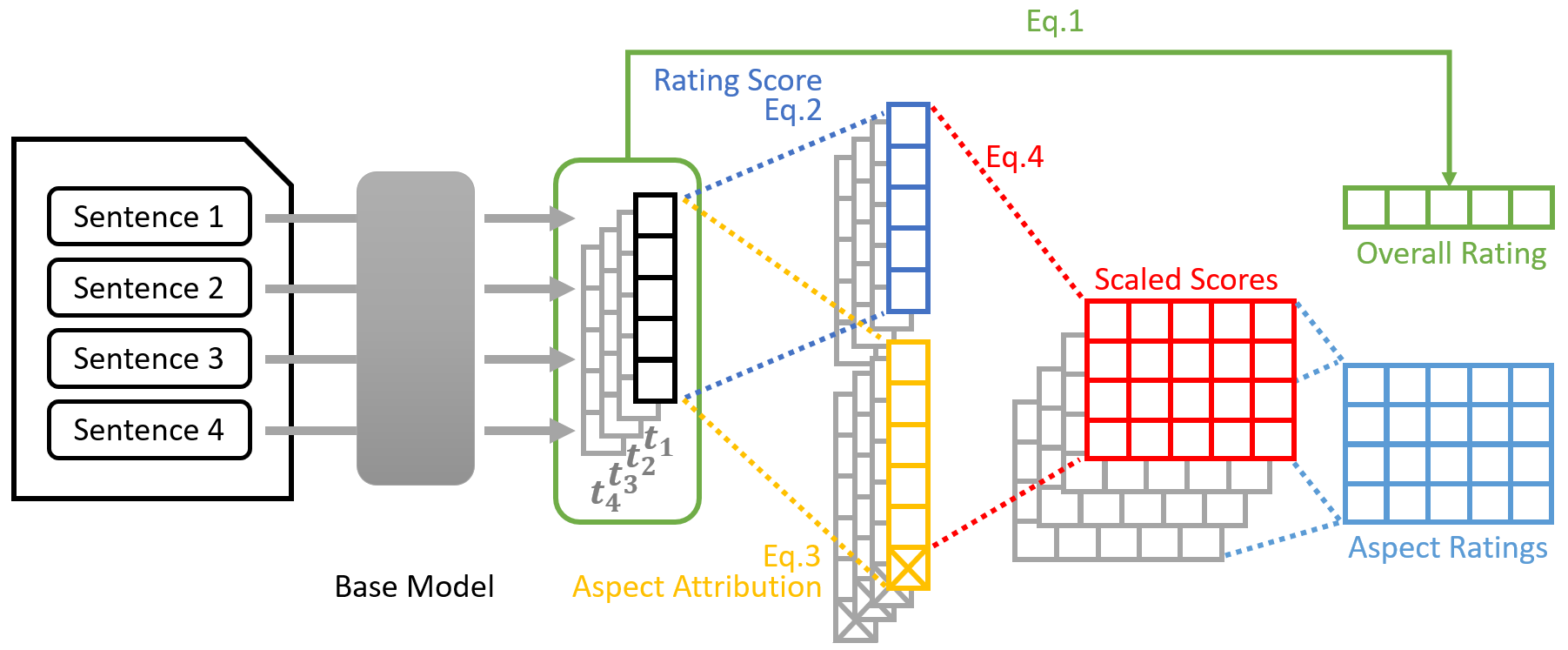}
  \caption{Architecture of SAAM Classification - 1}
  \label{fig:C1}
\end{figure}

The first variation of the SAAM classification model estimates the overall rating class using all features generated from all sentences by the convolution layer or the GRU cell directly. Each sentence's features are also passed into a fully connected softmax layer to estimate the 5-class rating distribution of each sentence, correspondingly. There is one such layer for every sentence in an input $r$ while the weights are shared. We refer to these layers as \textit{Rating Score Layers}. Another set of weights are used to estimate the aspect distribution of each sentence. We refer to these layers as \textit{Aspect Attribution Layers}. The resulting aspect distributions at the \textit{Aspect Attribution Layers} are then used to scale the rating scores from the \textit{Rating Score Layer} of each sentence, such that sentences with a high probability of belonging to a specific aspect exert a stronger influence on the ultimate aspect rating distributions at the document level. All scaled rating scores are then summed up for each aspect to estimate the final rating class for each aspect. The structure of this SAAM variation, together with the underlying K-CNN base, is visualized in Fig.\ref{fig:C1}.

More formally, given any base model such as a CNN or GRU and an input document $r$, we should be able to generate vector representation $\bm{t}$ of dimension $d$ for each sentence of the document. SAAM utilizes these sentence-level feature vectors generated by base networks to estimate latent distributions and ultimately sentiments of the document. For CNNs, these sentence representations are usually generated using max-pooling the filter activations along the sentence length dimension. For RNNs, this embedding can be obtained by using the final output at the last token. While for Bert-based models, the sentence embedding is usually generated by averaging all outputs along the sequence dimension or using the outputs of the \texttt{[CLS]} token. Thus, for an input document with $\left|s\right|$ number of sentences, a matrix $\bm{u}$ of dimension $|s| \times d$ can be obtained. This process is illustrated on the left side of Fig.\ref{fig:C1}.

To obtain a probability distribution of overall rating label for the entire review, all of the features in $\bm{u}$ are then passed into a fully connected softmax layer, the \textit{Overall Rating Layer}. Weights that corresponding to overall rating are labeled with a superscript $o$. This operation is shown in Fig.\ref{fig:C1} where a green arrow is marked with Eq.\ref{eq:C1_L_Overall}.
\begin{align*}
\label{eq:C1_L_Overall} L_{overall} (r)=softmax(\bm{W}^o \cdotp \bm{u} + \bm{b}^o ) \numberthis
\end{align*}

Like we discussed in the beginning of this section, to estimate the rating distribution of other aspects and carry out LSAA, feature values extracted from each sentence are fed into a \textit{rating score layer} and an \textit{aspect attribution layer}. For sentence $s_i$, the rating scores (un-normalized distribution) of sentence $s_i$ over $|C|$ rating classes are calculated by
\begin{align*}
\label{eq:11} score\left(s_i\right)=\left(\bm{W}^a \bm{t}_i + \bm{b}^a \right) \numberthis \\
\textrm{where} \quad \bm{W}^a \in \R^{d \times |C|} \quad \textrm{and} \quad \bm{b}^a \in \R^{|C|}
\end{align*}
In the case of a 5 star rating scheme, the above $|C|$ would equal to 5. This operation is demonstrated in Fig.\ref{fig:C1} where \textit{rating score layer} of each sentence is shown in yellow, 4 of such layer are drawn.

On the other hand, regarding the \textit{aspect attribution layer}, for a review of total $|A|$ aspects, we actually calculate the aspect attribution for sentence $s_i$ over $|A|+1$ aspects. The reason for this additional last element in each vector of attribution distribution, which we referred to as \textit{attribution to other-aspect}, is designed to relax the restriction for the model to some extent. It essentially allows the attribution process to ignore rating scores of some sentences if it deems necessary. Empirically, this structure does make the optimization process faster and allow the models to give a better result.
\begin{align*}
\label{eq:12} aspect(s_i) = softmax(\bm{W}^r \bm{t}_i + \bm{b}^r ) \numberthis \\
\textrm{where} \quad \bm{W}^r \in \R^{d \times (|A|+1)} \quad \textrm{and} \quad \bm{b}^r \in \R^{(|A|+1)}
\end{align*}
Notice in Eq.\ref{eq:11} and Eq.\ref{eq:12}, the same $\bm{W}^a$ and $\bm{W}^r$ are shared across all sentences. Four \textit{aspect attribution layers} are shown in Fig.\ref{fig:C1} marked using blue color.

Here, computing $aspect\left(s_i\right)$ should result in a vector $\R^{\left|A\right|+1}$ with the first $\left|A\right|$ elements represents how strong the sentence $s_i$ is associated to each aspect. We use $aspect\left(s_i\right)_{\left[1:|A|\right]}$ to denote these first $|A|$ elements. On the other hand, the last element of each aspect attribution, denoted as $aspect\left(s_i\right)_{\left[\lvert A \rvert + 1\right]}$, will not be associated with any of the actual aspect. We refer to it as the \textit{attribution of other-aspect}. As we will later explain, this additional attribution dimension gives the model the flexibility to determine if some sentences does not belong to any of the given aspects.

The first $|A|$ elements of \textit{aspect attribution layer} then distribute output from \textit{rating score layer} into respective aspects. More specifically, the scaled score for aspect $j$ of sentence $s_i$ would be equivalent to
\begin{align*}
\label{eq:13} scaledScore\left(s_i\right)^j = aspect\left(s_i\right)_{\left[j\right]} \cdotp score(s_i) \numberthis
\end{align*}
In Fig.\ref{fig:C1} this process is marked red. Notice how $scaledScore\left(s_i\right)$ is also equivalent to an outer product of the previous two layers, resulting in an matrix of size $\R^{(|A|+1) \times |C|}$, where row $j$ of this matrix is $scaledScore\left(s_i\right)^j$.

Lastly, these scaled scores for all sentences in a review are summed up element-wise per aspect. A softmax is then applied to the resulting vector to determine the distribution over rating classes for each aspect of document $r$:
\begin{equation} \label{eq:14} L_{aspect}^j(r) = softmax\left( \sum_{i=1}^{|s|} scaledScore(s_i)^j \right) \numberthis \end{equation}
This is shown in the bottom right corner of Fig.\ref{fig:C1} marked using light blue. We recall that the $ L_{aspect} $ only contains the rating distribution of aspects --- it does not include overall rating distribution. Since rating distribution for overall is directly evaluated using all sentence features $ \bm{u} $ at the \textit{Overall Rating Layer}. Also, it is worth noting that the distribution $L^{|A|+1}_{aspect}(r)$ is not used for estimating any label of the input document; it is the result of \textit{attribution of other-aspect} and hence disregarded.

\subsection{SAAM Classification--2 (SAAM-C2)}
\label{C2}
The second variation of the classification model is very similar to the first one. The only difference in this case is that we do not use a separate weight $\bm{W}^o$ to directly estimate the overall rating distribution. Instead, overall rating is predicted in a similar manner to other aspects, utilizing the sentence aspect attribution process. More specifically, this means for each sentence $s_i$, an \textit{aspect attribution layer} of size $\left|A\right|+2$ is used instead
\begin{align*}
\label{eq:15}
aspect\left( s_i \right) = softmax \left( \bm{W}^r \bm{t}_i + \bm{b}^r \right) \numberthis \\
\textrm{where} \quad \bm{W}^r \in \R^{(d) \times \left(|A|+2\right)} \quad \textrm{and} \quad \bm{b}^r \in \R^{\left(|A|+2\right)}
\end{align*}

Naturally, to estimate the overall rating of a review, we use the $\left(|A|+1\right)$th element of attribution layer: $aspect\left(s_i\right)_{\left[\left|A\right|+1\right]}$ to scale sentence level rating scores towards overall rating. These scores are then summed together and normalized using a softmax operation similar to SAAM-C1.

The main advantage of this modification over SAAM-C1 is the significant reduction in the size of parameters as the original overall weight matrix $\bm{W}^0$ is large. Classification-2 can thus use less memory and potentially less prone to overfitting. Moreover, this scheme can estimate the latent aspect attribution towards the overall aspect, if such information is indeed a point of interest.

\subsection{SAAM Regression (SAAM-R)}
\label{R1}
Apart from the more traditional rating classification task, we also present a variation of the SAAM where output layers are changed to real value regression for rating scores instead, while retaining the sentiment-aspect attribution mechanism. In this setting, the 5-star rating distribution is translated to a real value in the range of 1 to 5.

The regression variation of the architecture is architecturally similar to the first version of the classification model. We still connect all the features from all sentences to the output layer for overall score. However, in this case the overall output of the network is no longer a distribution over the rating classes, but a score without non-linearity. In addition to that, the score is normalized using the sentence count of the corresponding document:
\begin{align*}
\label{eq:18}
L_{overall} (r)= \frac{(\bm{W}^o \bm{u} + b^o )}{|s|} \quad \textrm{where} \quad \bm{W}^o \in \R^{|s| \times d} \numberthis
\end{align*}

Similarly, for each sentence we have a scalar score
\begin{align*}
\label{eq:19}
score\left(s_i \right)=\bm{W}^a \bm{t}_i + b^a \numberthis
\end{align*}

On the other hand, \textit{aspect attribution layer} is kept the same as Classification-1 Eq.\ref{eq:12} in this paradigm. The sentence-level scalar score of sentence $s_i$ is then scaled by multiplying with aspect weights. So for aspect $j$, this is calculated by

\begin{equation} \label{eq:21} scaledScore \left( s_i \right)^j = aspect\left( s_i \right)_{\left[ j \right]} \times score\left( s_i \right) \end{equation}

This operation results in a total of $|A|$ scalar score for each sentence, with each value corresponding to one of the aspects. And the final score for aspect $j$ is calculated by

\begin{equation} \label{eq:22} L_{aspect}^j (r) = \frac{\sum_{i=1}^{|s|} scaledScore\left( s_i \right)^j }{\sum_{i=1}^{|s|} aspect\left( s_i \right)_{[j]} } \end{equation}

Notice the regression scores for aspects are normalized differently compared to the overall score as shown in Eq.\ref{eq:18}: the scoring for each aspect is normalized with the total probability assigned to that aspect by the attribution layer, instead of the number of sentences in the corresponding review. This normalization makes the aspect scoring process equivalent to a weighted average of sentence aspect scoring, with attribution distribution being the weights.

This difference in normalization is due to the overall score being designed to be an average of sentence scores - it would be problematic if a longer review with a high number of positive sentences goes above the 1-5 score range - assuming the padding sentences getting scores close to 0. On the other hand, attribution layer has the capability to ``throw away'' scores from the padding sentences when calculating the aspect scores by assigning them a 100\% weight on \textit{attribution of other-aspect}, i.e. $ aspect\left( s_i \right)_{\left[ |A|+1 \right]}$. Hence the sentence count normalization is no longer needed.

\subsection{Intuitions}

\begin{figure}[t!]
    \includegraphics[width=\linewidth]{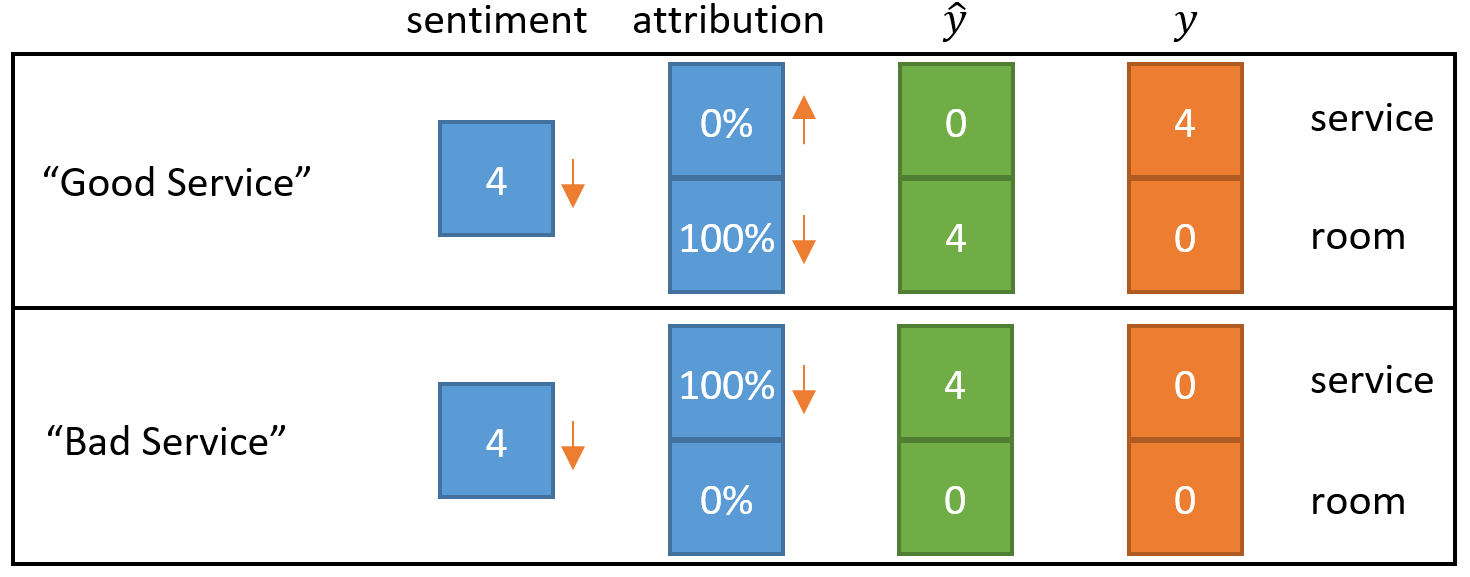}
  \caption{Optimization of attribution layer}
  \label{fig:opti}
\end{figure}

We provide a simplified example regarding how latent attribution can discover the correct aspect given enough examples. Fig.\ref{fig:opti} shows two training documents, each containing only one sentence, being processed using a simplified SAAM-R model. We picked the regression model and only two aspects, Service and Room, for easier demonstration. However, the idea should be able to generalize to other variants as well as more aspects.

Blue boxes are internal parameters produced by SAAM's Rating Scoring Layer and Attribution Layer (marked as \textit{sentiment} and \textit{attribution} respectively). Green boxes are output values resulting from the element-wise product of the former two layers. Lastly, orange boxes are ground truth scores.

Recall that for SAAM-Regression, the document-level aspect rating predictions are calculated by

\begin{equation}  \hat{y} = sentiment \otimes attribution \end{equation}

and the loss can be expressed simply as: $ \label{eq:loss} loss = (\hat{y} - y) ^ 2 $ .

For the first sentence ``Good Service'', assuming the sentiment layer produces the correct score but the attribution layer is wrong by attributing all the sentiment scores into room aspect. When optimizing the sentiment layer and attribution layer using gradient descend, the gradients' directions are indicated using orange arrows for each blue value. As readers can see, the attribution layer will be slightly adjusted towards the correct attribution, which is service 100\% and room 0\%. 

In the second example, ``Bad Service'', assuming the attribution layer this time produces the correct aspect distribution, but the sentiment layer mistakenly produced a very high sentiment score. In this case, the gradient descent will pass through the attribution layer and decrease the sentiment score.

Although, in both examples, the gradient descent process has produced some side effects: the sentiment layer in the first case and attribution layer in the second case were optimized in the wrong direction. But ultimately given enough examples and training steps the system should converge correctly. Imagine a third sentence ``Good Room'' with correct ground truth, in which circumstances the optimization process should have only one possible solution in the blue boxes.

\section{Experiments and Evaluations}
\label{experiment}

\subsection{Data}
\label{data}
We use the TripAdvisor hotel review dataset from \cite{wang2010latent} and BeerAdvocate data previously used in works such as \cite{mcauley2012learning} to examine the performance of our framework.

The TripAdvisor dataset consists of 108,891 reviews across 1,850 hotels. Each hotel review in the original raw data is associated with one overall rating and five aspect ratings - Value, Room, Location, Cleanliness, Service. For our experiment, only reviews with more than three sentences, and all of the five aspects rated were selected. Out of the 14,906 reviews that meet the above requirements, 75\%, 25\% of the documents were selected as training and testing sets, respectively. 1000 reviews were picked from the training set as a development set for tuning hyper-parameters. After determined the parameters, the models were re-trained using all training samples available.

Similar parsing and selection process was also applied to the BeerAdvocate dataset, and 100,000 beer reviews were selected for our experiment. Aspects associated with each beer review include Appearance, Taste, Palate, and Aroma. Among these, aspect Palate can be roughly understood as ``mouthfeel''. The advantage of this dataset is that its aspects are more independent of each other --- whereas, in the case of hotel reviews, value, room, and cleanliness are often strongly correlated. This property of the BeerAdvocate dataset allows us to better evaluate the sentence-level aspect attribution process's correctness. Similar to the TripAdvisor dataset, 75\%, 25\% of the documents were selected as training and testing set, respectively, and 1,000 reviews were picked as a development set for tuning model parameters.

We will use the TripAdvisor hotel review dataset to evaluate our classification modules SAAM-C1 and SAAM-C2, while using the BeerAdvocate beer review dataset to evaluate our regression variant of the module SAAM-R. This is due to both this paper's space constraint as well as the fact that BeerAdvocate review ratings are in 0.5 increments.

\subsection{Evaluation of Document-Level MASA}
Because our proposed SAAM is an add-on module that can be combined with almost all modern neural network architectures, the document level sentiment analysis performance of the complete model (base + SAAM) is determined by both components. In the following experiments, we opt to use two representative models as base models to better highlight the characteristics of SAAM.

The first base model is K-CNN proposed in \cite{kim2014convolutional}. In its original form, the model uses a total of 300 convolutional filters (100 of each size) to extract features from each review. Then a fully connected softmax is applied on top of that to estimate the label probability distribution. We trained a separate model for each aspect of the reviews as baselines. We then replace the fully connected layers at the end of K-CNN with our SAAM to demonstrate that we can improve the performance of the overall model.

Furthermore, we have also included a version of the CNN, which we refer to as Expanded CNN (E-CNN), to demonstrate that the performance improvement we observed from using SAAM is not merely due to an increase in the number of parameters. Specifically, in this baseline, reviews are also divided into sentences. Each sentence is then passed to the CNN layer to generate 300 dimension embedding. All of the features generated from sentences are then concatenated and passed to a fully connected softmax layer for classification. Notice this is similar to how the overall rating is estimated in the SAAM-C1 scheme (\ref{C1}) we proposed, shown in Eq.\ref{eq:C1_L_Overall}. A total of $|A|$ such fully connected softmax layers in the model to concurrently train and estimate all aspects.

The second base model we chose is a GRU based RNN \cite{cho2014learning}. Similar to CNN, we set the hidden state vector to 300 and trained a separate model for each aspect of the dataset as baselines. We then replace the final fully-connected layers with our SAAM to demonstrate its flexibility and performance improvement over the base models. 

We have also included 3 classification baselines Hierarchical LSTM \cite{ruder2016hierarchical}, Doc2Vec \cite{le2014distributed} and SVM \cite{joachims1999making} for TripAdvisor hotel review classification task, and 2 regression baselines Linear Regression and SVM regression \cite{joachims1999making} for BeerAdvocate beer review regression task. We included these referencing baselines to help readers interpret the difficulty of our task and dataset and use them as a benchmark for estimating the expressiveness of our proposed modules. The Hierarchical LSTM proposed in \cite{ruder2016hierarchical} consists of two levels of LSTM networks: one working at the word level to generate sentence embedding vectors, another takes these sentence embedding vectors as input and estimates sentiment polarity for each sentence in a document. To adapt this model as one of our baselines, we modify the model by concatenating the last output vectors from the sentence-level bi-directional LSTM and feeding the resulting vector to several dense layers, where each layer corresponds to one of the aspects.

We also note that the additional computational time required to train SAAM is not significant, as the additional parameter matrix $W^a$ and $W^r$ is relatively small. We tested all SAAM variants on one Nvidia Titan RTX GPU; the increase in training time was around 10\% to 30\% longer compared to base model CNN and RNN.

\subsection{MASA Results}
\label{hotel result}

\begin{table*}[]
	\centering
	\resizebox{\textwidth}{!}{%
		\begin{tabular}{lllllllllllllll}
			\thickhline
			& \multicolumn{2}{c}{}                                                 & \multicolumn{2}{c}{Aspect 1}                                           & \multicolumn{2}{c}{Aspect 2}                                           & \multicolumn{2}{c}{Aspect 3}                                           & \multicolumn{2}{c}{Aspect 4}                                           & \multicolumn{2}{c}{Aspect 5}                                           & Avg                                      & Avg                           \\ 
			& \multicolumn{2}{c}{Overall}                                          & \multicolumn{2}{c}{Value}                                              & \multicolumn{2}{c}{Room}                                               & \multicolumn{2}{c}{Location}                                           & \multicolumn{2}{c}{Cleanliness}                                        & \multicolumn{2}{c}{Service}                                            & \multicolumn{2}{l}{}                                                    \\  
			\multirow{-3}{*}{} & Acc                                   & MSE                           & Acc                                   & MSE                                    & Acc                                   & MSE                                    & Acc                                   & MSE                                    & Acc                                   & MSE                                    & Acc                                   & MSE                                    & Acc.                                   & MSE                                    \\
			\hline
			
			K-CNN            & \cellcolor[HTML]{C0C0C0}58.0          & 0.715                         & \cellcolor[HTML]{C0C0C0}50.8          & 0.943                                  & \cellcolor[HTML]{C0C0C0}45.1          & 1.061                                  & \cellcolor[HTML]{C0C0C0}44.8          & 1.302                                  & \cellcolor[HTML]{C0C0C0}47.5          & 0.995                                  & \cellcolor[HTML]{C0C0C0}50.3          & 1.319                                  & \cellcolor[HTML]{C0C0C0}47.70          & 1.124                                  \\
			E-CNN            & \cellcolor[HTML]{C0C0C0}58.6          & 0.600                         & \cellcolor[HTML]{C0C0C0}49.9          & 0.883                                  & \cellcolor[HTML]{C0C0C0}41.8          & 1.135                                  & \cellcolor[HTML]{C0C0C0}42.9          & 1.107                                  & \cellcolor[HTML]{C0C0C0}46.1          & 1.076                                  & \cellcolor[HTML]{C0C0C0}48.6          & 1.224                                  & \cellcolor[HTML]{C0C0C0}45.86          & 1.085                                  \\
			
			CNN+SAAM-C1         & \cellcolor[HTML]{C0C0C0}58.3 & 0.706 & \cellcolor[HTML]{C0C0C0}\textbf{51.6} & 0.888 & \cellcolor[HTML]{C0C0C0}\textbf{47.2} & \textbf{0.985} & \cellcolor[HTML]{C0C0C0}44.7 & 1.308 & \cellcolor[HTML]{C0C0C0}\textbf{50.2} & 1.042 & \cellcolor[HTML]{C0C0C0}\textbf{51.6} & \textbf{1.138} & \cellcolor[HTML]{C0C0C0}\textbf{49.06} & 1.072 \\
			CNN+SAAM-C2         & \cellcolor[HTML]{C0C0C0}58.0 & 0.62 & \cellcolor[HTML]{C0C0C0}\textbf{51.8} & \textbf{0.803} & \cellcolor[HTML]{C0C0C0}\textbf{48.2} & \textbf{0.906} & \cellcolor[HTML]{C0C0C0}\textbf{45.3} & 1.166 & \cellcolor[HTML]{C0C0C0}\textbf{49.3} & \textbf{0.927} & \cellcolor[HTML]{C0C0C0}\textbf{51.0} & \textbf{1.039} & \cellcolor[HTML]{C0C0C0}\textbf{49.12} & \textbf{0.968} \\
			
			\hline

			RNN            & \cellcolor[HTML]{C0C0C0}58.2          & 0.647                         & \cellcolor[HTML]{C0C0C0}51.4          & 0.891                                  & \cellcolor[HTML]{C0C0C0}44.9          & 1.158                                  & \cellcolor[HTML]{C0C0C0}43.5          & 1.467                                  & \cellcolor[HTML]{C0C0C0}45.9          & 1.214                                  & \cellcolor[HTML]{C0C0C0}48.4          & 1.209                                  & \cellcolor[HTML]{C0C0C0}48.72          & 1.098                                  \\
			
			RNN+SAAM-C1         & \cellcolor[HTML]{C0C0C0}56.6  & 0.722 & \cellcolor[HTML]{C0C0C0}\textbf{54.9} & \textbf{0.772} & \cellcolor[HTML]{C0C0C0}\textbf{49.0} & \textbf{0.976} & \cellcolor[HTML]{C0C0C0}\textbf{45.8} & 1.407 & \cellcolor[HTML]{C0C0C0}\textbf{49.8} & \textbf{1.041} & \cellcolor[HTML]{C0C0C0}\textbf{51.5} & \textbf{1.100} & \cellcolor[HTML]{C0C0C0}\textbf{51.27} & 1.003 \\
			RNN+SAAM-C2         & \cellcolor[HTML]{C0C0C0}\textbf{60.2} & \textbf{0.625}  & \cellcolor[HTML]{C0C0C0}\textbf{54.1} & \textbf{0.824}  & \cellcolor[HTML]{C0C0C0}\textbf{49.5} & \textbf{0.969} & \cellcolor[HTML]{C0C0C0}\textbf{46.6} & \textbf{1.279} & \cellcolor[HTML]{C0C0C0}\textbf{50.4} & \textbf{1.021} & \cellcolor[HTML]{C0C0C0}\textbf{52.3} & \textbf{1.052} & \cellcolor[HTML]{C0C0C0}\textbf{52.19} & \textbf{0.962} \\

            \hline
			
            Hi-LSTM         & \cellcolor[HTML]{C0C0C0}61.6 & 0.533                & \cellcolor[HTML]{C0C0C0}54.7          & 0.751                                  & \cellcolor[HTML]{C0C0C0}46.4          & 1.029                                  & \cellcolor[HTML]{C0C0C0}44.8          & 1.216                                  & \cellcolor[HTML]{C0C0C0}47.1          & 1.052                                  & \cellcolor[HTML]{C0C0C0}48.7          & 1.234                                  & \cellcolor[HTML]{C0C0C0}50.5           & 0.969                                  \\
			Doc2Vec          & \cellcolor[HTML]{C0C0C0}54.1          & 0.829                         & \cellcolor[HTML]{C0C0C0}47.8          & 1.087                                  & \cellcolor[HTML]{C0C0C0}42.3          & 1.305                                  & \cellcolor[HTML]{C0C0C0}44.7          & 1.439                                  & \cellcolor[HTML]{C0C0C0}45.1          & 1.291                                  & \cellcolor[HTML]{C0C0C0}47.3          & 1.585                                  & \cellcolor[HTML]{C0C0C0}45.44          & 1.341                                  \\
			SVM              & \cellcolor[HTML]{C0C0C0}29.2          & 1.892                         & \cellcolor[HTML]{C0C0C0}35.5          & 2.368                                  & \cellcolor[HTML]{C0C0C0}33.9          & 2.368                                  & \cellcolor[HTML]{C0C0C0}8.4           & 9.010                                  & \cellcolor[HTML]{C0C0C0}32.5          & 1.917                                  & \cellcolor[HTML]{C0C0C0}33.3          & 2.375                                  & \cellcolor[HTML]{C0C0C0}28.72          & 3.608                                  \\

			\thickhline
		\end{tabular}
	}
	\caption{Performance of proposed SAAM classification variants against corresponding base models and other baselines, experimented on TripAdvisor hotel review dataset.}
	\label{table:hotel result}
\end{table*}

Table \ref{table:hotel result} shows the results of classification variants SAAM-C1, SAAM-C2 on top of CNN and GRU-RNN compared against the base version of these two models, evaluated on TripAdvisor testing set. Both prediction Accuracy (Acc) and Mean Squared Error (MSE) are calculated for overall and other five aspects. To calculate MSE, the predicted classes are regarded as real values when calculating. On the right-most two columns, averaged five aspect Accuracy and MSE are calculated for easier comparison. We use bold texts to highlight statistically significant performance improvement of SAAM applied models over their corresponding base models.

From Table \ref{table:hotel result}, we note that our proposed SAAM-C1 and SAAM-C2 models can provide a consistent performance improvement over their corresponding base model counterparts. More specifically, stacking SAAM-C1 and SAAM-C2 on top of CNN and RNN on average improves the aspect sentiment classification accuracy by 2 to 3 percent. In certain aspects such as \textit{Room} and \textit{Cleanliness}, the improvements in accuracy are as much as 5 percent. We also note that there is little to no improvements on the \textit{Overall} rating classification. One reason for this could be that overall sentiment classification is relatively easy, as the model does not need to learn aspect specific feature combinations, and reviewer behavior is more consistent for overall ratings.

As a reference, Hi-LSTM provides an additional 1 to 4 percent of accuracy when comparing with base versions of RNN. This is likely due to the additional expressiveness offered by the second layer of LSTM, which can selectively pass through sentence-level features to document level output to allow more accurate distribution estimation. In other words, the performance advantage of Hi-LSTM can be attributed to its more expressive sentence-to-document connection. As a comparison, after combining SAAM-C1 and C2 with the RNN base model, the performance gaps between RNN and Hi-LSTM have been eliminated and, in some cases, reversed, indicating that SAAM significantly improves the expressiveness and information flow from sentence level to document level.

\begin{table*}[]
	\centering
	\resizebox{\textwidth}{!}{%
		\begin{tabular}{lllllllllllll}
			\thickhline
			
			\multirow{2}{*}{} & \multicolumn{2}{c}{Overall}    & \multicolumn{2}{c}{\begin{tabular}[c]{@{}c@{}}Aspect 1\\ Appearance\end{tabular}} & \multicolumn{2}{c}{\begin{tabular}[c]{@{}c@{}}Aspect 2\\ Taste\end{tabular}} & \multicolumn{2}{c}{\begin{tabular}[c]{@{}c@{}}Aspect 3\\ Palate\end{tabular}} & \multicolumn{2}{c}{\begin{tabular}[c]{@{}c@{}}Aspect 4\\ Aroma\end{tabular}} & \multicolumn{2}{c}{Average}      \\ 
			         & MSE            & R2             & MSE            & R2             & MSE            & R2             & MSE            & R2             & MSE            & R2             & MSE            & R2             \\
			\hline
			
			E-CNN     & 0.267 & 0.423 & 0.228 & 0.325 & 0.260  & 0.454 & 0.239 & 0.425 & 0.258 & 0.408 & 0.246 & 0.403   \\
			
			CNN+SAAM-R   & 0.264 & 0.431 & \textbf{0.208} & \textbf{0.386} & \textbf{0.207} & \textbf{0.564} & 0.220 & 0.471 & \textbf{0.219} & \textbf{0.498} & \textbf{0.213} & \textbf{0.480} \\ 
			
			\hline
			
			RNN  & 0.256 & 0.448  & 0.209 & 0.383 & 0.231 & 0.514 & 0.237 & 0.429 & 0.243 & 0.445 & 0.235 & 0.443  \\

			RNN+SAAM-R   & \textbf{0.228} & \textbf{0.508} & 0.195 & 0.424 & \textbf{0.182} & \textbf{0.617} & \textbf{0.202} & \textbf{0.514} & \textbf{0.199} & \textbf{0.542} & \textbf{0.201} & \textbf{0.521}          \\ 
			
            \hline

			Linear Regr & 0.307          & 0.338          & 0.255          & 0.246          & 0.266          & 0.440          & 0.287          & 0.308          & 0.285          & 0.346          & 0.273          & 0.335          \\
			SVM         & 0.272          & 0.414          & 0.226          & 0.332          & 0.235          & 0.505          & 0.253          & 0.391          & 0.252          & 0.421          & 0.242          & 0.412          \\
			\thickhline
		\end{tabular}
	}
	\caption{Performance of proposed SAAM regression variants against corresponding base models and other baselines, experimented on BeerAdvocate beer review dataset.}
	\label{table:beer result}
\end{table*}

Table \ref{table:beer result} shows the performance of our SAAM regression models (SAAM-R) based on CNN and RNN, as well as base models and other referencing baselines evaluated on BeerAdvocate beer review data testing set. We use bold texts to highlight statistically significant performance improvement of SAAM applied models over their corresponding base models. Once again, we can see by adding our SAAM regression module on top of the base model, we can significantly reduce the error when comparing to base models. Among all aspects, we observed that base models CNN and RNN have relatively poor performance on aspect \textit{Taste} and \textit{Aroma}, which may be caused by languages describing these two aspects being very similar. The attribution mechanism in our model can alleviate this issue by redirecting the latent sentence-level sentiment to the correct aspect.

\subsection{Evaluation of Latent Sentence-Level Aspect Attribution}

One of our SAAM framework's key advantages is that it can leverage the latent aspect attributed to each sentence and organically combine them. In this section, we evaluate the LSAA facet of our models. We let two human labelers manually label 1,000 sentences with aspects in each of the datasets. For both of the dataset, the set of possible labels included names of the existing aspects and an additional label ``\textit{none}'' --- which indicates the labeler thinks the sentence is not related to any of the aspects. The labeling from two labelers achieved a Cohen's Kappa agreement score of 0.66, indicating significant agreement but not perfect. On the other hand, the beer review dataset shows a better agreement score of 0.70, reinforcing our observation that aspects in beer review are more independent and unambiguous.

In addition to the human labelers, we take advantage of the review format many reviewers follow in the BeerAdvocate dataset as another set of ground truth. More specifically, many reviewers on BeerAdvocate use ``A:'', ``S:'', ``M:'' and ``T:'' to signify the beginning of corresponding review segments\footnote{``A'' for Appearance; ``S'' for Smell, corresponding to Aroma aspect; ``M'' for Mouthfeel, corresponding to Palate aspect; ``T'' for Taste.}. We selected around 16,000 sentences that have these prefixes and marked them with corresponding correct labels.

\begin{table}[]
	\centering

	\begin{tabular}{lccccc}
		\hline
		\hline
		  & Hotel 1 & Hotel 2 & Beer 1 & Beer 2 & Beer Keywd \\
		  CNN+C1 & 0.32 & 0.35 & - & - & - \\
		  CNN+C2 & 0.48 & 0.47 & - & - & - \\
		  CNN+R  & -    & -    & 0.63 & 0.61 & 0.87 \\
		\hline
		  GRU+C1 & 0.46 & 0.50 & - & - & - \\
		  GRU+C2 & 0.55 & 0.52 & - & - & - \\
		  GRU+R  & -    & -    & 0.68 & 0.64 & 0.95 \\
		\hline
		\hline
	\end{tabular}

	\caption{Evaluation of our SAAM framework's ability to estimate latent sentence aspects. Accuracy is reported against two independent human labelings on both datasets and a keyword based labeling method on BeerAdvocate dataset.}
	\label{table:aspect}
\end{table}

To obtain the sentence-level latent aspects attributed by SAAM, we looked at the estimated latent aspect distribution ($aspect\left( s_i \right)$) for that sentence. If the dominant value of the learned aspect distribution dovetailed with the aspect labeled by the labeler, it is considered as a correct attribution. Table \ref{table:aspect} shows these evaluation results of SAAM-C1, SAAM-C2 on the hotel review dataset, and SAAM-R on the beer review dataset. We can observe that almost all model combinations can attribute sentences to aspects at reasonably high accuracy. Among these, the regression model based on GRU is the best performer on this task. Moreover, we can see the regression models demonstrate even stronger agreement with the keyword-based labeling. This means the models have successfully learned these keywords and are using them as strong signals when conducting latent aspect attribution.

It is worth noting that due to inherent overlapping between aspect categories, reviewer subjectivity, and vague nature of some of the aspects, this LSAA task is non-trivial. Considering it is a latent variable and there are 4 to 5 potential classes, the above results indicate good performance. 

\subsection{Snippet Extraction}

In addition to estimating latent aspect distribution, SAAM can also estimate the latent sentiment distribution ($score\left( s_i \right)$) for each sentence. We believe there is much exciting opportunity for information extraction by combining this latent information discovered through LSAA. This section demonstrates one interesting possible application of review snippet extraction per aspect, inspired by several existing review summarization work such as \cite{li2010structure}. Particularly, it is interesting for those cases where the overall review rating is positive while one aspect is evaluated negatively (or vice-versa) and whether our model is able to explain the discrepancy. Here, we show some qualitative results using SAAM-R to provide an intuitive understanding of this application and SAAM framework.

\vspace{0.3cm}

\noindent
\textbf{Review 1, Overall 5 Stars:} ``\textit{spent 5 days at excellence at Punta Cana, most of the people who work at the hotel were very pleasant \ldots }'' \\
Sentiment snippet for Service aspect via the lowest sentiment score: 
\begin{itemize}
\item \textit{``internet service was not available in the room and barely in the lobby area''} \textbf{[Service, -2.89]}
\end{itemize}

\vspace{0.3cm}

\noindent
\textbf{Review 2, Overall 1 star:} ``\textit{I do not know where to start. the roaches in the room, the rude waiters, bartenders, front desk, the dead flies that stayed on our friends' mirror the entire stay, the average at best food \ldots }'' \\
Sentiment snippet for Location and Cleanliness aspects via the highest sentence score:
\begin{itemize}
\item \textit{``the beach was fabulous''} \textbf{[Location, 5.99]}
\item \textit{``the resort itself, décor, pool, beach access was great''} \textbf{[Cleanliness, 5.90]}
\end{itemize}

\vspace{0.3cm}

\noindent
\textbf{Review 3, Overall 5 stars with 3 stars in Location aspect:} ``\textit{Was awesome. my wife and I traveled to excellence 11/20-11/26 and had a great time \ldots}'' \\
Sentiment snippet for Location aspect via the lowest sentence score:
\begin{itemize}
\item \textit{``the worst part about this resort is the drive there and back, the roads are terrible and it is over an hour''} \textbf{[Location, -1.53]}
\end{itemize}

\vspace{0.3cm}

\noindent
\textbf{Review 4, Overall 4 stars with 2.5 stars in Palate aspect:} ``\textit{A: Pours a clear yellow with a mild white head, good retention \ldots}'' \\
Sentiment snippet by extracting the only Palate sentence:
\begin{itemize}
\item \textit{``M: Very light-bodied, watery, light base beer for sure.''} \textbf{[Palate, -0.96]}
\end{itemize}

\section{Conclusion}
\label{conclusion}
In this paper, we presented a novel add-on framework called the sentiment-aspect attribution module (SAAM) that can be combined with common deep learning architectures to solve the problem of multi-aspect sentiment analysis. The proposed SAAM addresses the token-to-doc connection bottleneck problem using an intuitive and expressive latent sentiment-aspect attribution (LSAA) process. Furthermore, the LSAA process also facilitates fine-grained sentiment analysis and summarization. Two classification and one regression variants of the SAAM were demonstrated and tested on both CNN and RNN based networks. Experimental results on real-world hotel review dataset and beer review dataset demonstrated significant performance improvement over original base networks. Lastly, we also demonstrated the potential of using sentence level latent features generated by SAAM for aspect-specific or sentiment-specific snippet extraction. We understand there is a lot of room for improvement for this iteration of SAAM. However, we believe this work presents a fascinating new angle in solving multi-labelled document classification problems.

\vspace{0.3cm}
\noindent\textbf{Acknowledgement}
\noindent\textit{Research was supported in part by grants NSF 1838147 and ARO W911NF-20-1-0254. The views and conclusions contained in this document are those of the authors and not of the sponsors.}



\bibliographystyle{IEEEtran}
%
\bibliography{mybib}

\begin{thebibliography}{10}
\providecommand{\url}[1]{#1}
\csname url@samestyle\endcsname
\providecommand{\newblock}{\relax}
\providecommand{\bibinfo}[2]{#2}
\providecommand{\BIBentrySTDinterwordspacing}{\spaceskip=0pt\relax}
\providecommand{\BIBentryALTinterwordstretchfactor}{4}
\providecommand{\BIBentryALTinterwordspacing}{\spaceskip=\fontdimen2\font plus
\BIBentryALTinterwordstretchfactor\fontdimen3\font minus
  \fontdimen4\font\relax}
\providecommand{\BIBforeignlanguage}[2]{{%
\expandafter\ifx\csname l@#1\endcsname\relax
\typeout{** WARNING: IEEEtran.bst: No hyphenation pattern has been}%
\typeout{** loaded for the language `#1'. Using the pattern for}%
\typeout{** the default language instead.}%
\else
\language=\csname l@#1\endcsname
\fi
#2}}
\providecommand{\BIBdecl}{\relax}
\BIBdecl

\bibitem{kim2014convolutional}
Y.~Kim, ``Convolutional neural networks for sentence classification,''
  \emph{arXiv preprint arXiv:1408.5882}, 2014.

\bibitem{wang2010latent}
H.~Wang, Y.~Lu, and C.~Zhai, ``Latent aspect rating analysis on review text
  data: a rating regression approach,'' in \emph{Proceedings of the 16th ACM
  SIGKDD international conference on Knowledge discovery and data
  mining}.\hskip 1em plus 0.5em minus 0.4em\relax ACm, 2010, pp. 783--792.

\bibitem{wang2011latent}
------, ``Latent aspect rating analysis without aspect keyword supervision,''
  in \emph{Proceedings of the 17th ACM SIGKDD international conference on
  Knowledge discovery and data mining}, 2011, pp. 618--626.

\bibitem{mcauliffe2008supervised}
J.~D. Mcauliffe and D.~M. Blei, ``Supervised topic models,'' in \emph{Advances
  in neural information processing systems}, 2008, pp. 121--128.

\bibitem{titov2008joint}
I.~Titov and R.~McDonald, ``A joint model of text and aspect ratings for
  sentiment summarization,'' in \emph{proceedings of ACL-08: HLT}, 2008, pp.
  308--316.

\bibitem{lu2011multi}
B.~Lu, M.~Ott, C.~Cardie, and B.~K. Tsou, ``Multi-aspect sentiment analysis
  with topic models,'' in \emph{2011 IEEE 11th international conference on data
  mining workshops}.\hskip 1em plus 0.5em minus 0.4em\relax IEEE, 2011, pp.
  81--88.

\bibitem{yu2011aspect}
J.~Yu, Z.-J. Zha, M.~Wang, and T.-S. Chua, ``Aspect ranking: identifying
  important product aspects from online consumer reviews,'' in
  \emph{Proceedings of the 49th Annual Meeting of the Association for
  Computational Linguistics: Human Language Technologies-Volume 1}.\hskip 1em
  plus 0.5em minus 0.4em\relax Association for Computational Linguistics, 2011,
  pp. 1496--1505.

\bibitem{snyder2007multiple}
B.~Snyder and R.~Barzilay, ``Multiple aspect ranking using the good grief
  algorithm,'' in \emph{Human Language Technologies 2007: The Conference of the
  North American Chapter of the Association for Computational Linguistics;
  Proceedings of the Main Conference}, 2007, pp. 300--307.

\bibitem{mcauley2012learning}
J.~McAuley, J.~Leskovec, and D.~Jurafsky, ``Learning attitudes and attributes
  from multi-aspect reviews,'' in \emph{2012 IEEE 12th International Conference
  on Data Mining}.\hskip 1em plus 0.5em minus 0.4em\relax IEEE, 2012, pp.
  1020--1025.

\bibitem{pontiki2016semeval}
\BIBentryALTinterwordspacing
M.~Pontiki, D.~Galanis, H.~Papageorgiou, I.~Androutsopoulos, S.~Manandhar,
  M.~AL-Smadi, M.~Al-Ayyoub, Y.~Zhao, B.~Qin, O.~{De Clercq}, V.~Hoste,
  M.~Apidianaki, X.~Tannier, N.~Loukachevitch, E.~Kotelnikov, N.~Bel, S.~M.
  Jim{\'{e}}nez-Zafra, and G.~Eryiğit, ``{SemEval-2016 Task 5: Aspect Based
  Sentiment Analysis},'' in \emph{Proceedings of the 10th International
  Workshop on Semantic Evaluation (SemEval-2016)}.\hskip 1em plus 0.5em minus
  0.4em\relax Stroudsburg, PA, USA: Association for Computational Linguistics,
  2016, pp. 19--30. [Online]. Available:
  \url{http://aclweb.org/anthology/S16-1002}
\BIBentrySTDinterwordspacing

\bibitem{tang2016effective}
\BIBentryALTinterwordspacing
D.~Tang, B.~Qin, X.~Feng, and T.~Liu, ``Effective {LSTM}s for target-dependent
  sentiment classification,'' in \emph{Proceedings of {COLING} 2016, the 26th
  International Conference on Computational Linguistics: Technical
  Papers}.\hskip 1em plus 0.5em minus 0.4em\relax Osaka, Japan: The COLING 2016
  Organizing Committee, Dec. 2016, pp. 3298--3307. [Online]. Available:
  \url{https://www.aclweb.org/anthology/C16-1311}
\BIBentrySTDinterwordspacing

\bibitem{ruder2016hierarchical}
\BIBentryALTinterwordspacing
S.~Ruder, P.~Ghaffari, and J.~G. Breslin, ``{A Hierarchical Model of Reviews
  for Aspect-based Sentiment Analysis},'' in \emph{Proceedings of the 2016
  Conference on Empirical Methods in Natural Language Processing}.\hskip 1em
  plus 0.5em minus 0.4em\relax Stroudsburg, PA, USA: Association for
  Computational Linguistics, 2016, pp. 999--1005. [Online]. Available:
  \url{http://aclweb.org/anthology/D16-1103}
\BIBentrySTDinterwordspacing

\bibitem{yang-etal-2016-leveraging}
\BIBentryALTinterwordspacing
F.~Yang, A.~Mukherjee, and Y.~Zhang, ``Leveraging multiple domains for
  sentiment classification,'' in \emph{Proceedings of {COLING} 2016, the 26th
  International Conference on Computational Linguistics: Technical
  Papers}.\hskip 1em plus 0.5em minus 0.4em\relax Osaka, Japan: The COLING 2016
  Organizing Committee, Dec. 2016, pp. 2978--2988. [Online]. Available:
  \url{https://www.aclweb.org/anthology/C16-1280}
\BIBentrySTDinterwordspacing

\bibitem{boumber2018experiments}
D.~Boumber, Y.~Zhang, and A.~Mukherjee, ``Experiments with convolutional neural
  networks for multi-label authorship attribution,'' in \emph{Proceedings of
  the Eleventh International Conference on Language Resources and Evaluation
  (LREC 2018)}, 2018.

\bibitem{zhang2020birds}
Y.~Zhang, F.~Yang, Y.~Zhang, E.~Dragut, and A.~Mukherjee, ``Birds of a feather
  flock together: Satirical news detection via language model
  differentiation,'' 2020.

\bibitem{bahdanau2016neural}
D.~Bahdanau, K.~Cho, and Y.~Bengio, ``Neural machine translation by jointly
  learning to align and translate,'' 2016.

\bibitem{devlin2019bert}
J.~Devlin, M.-W. Chang, K.~Lee, and K.~Toutanova, ``Bert: Pre-training of deep
  bidirectional transformers for language understanding,'' 2019.

\bibitem{xu-etal-2019-bert}
\BIBentryALTinterwordspacing
H.~Xu, B.~Liu, L.~Shu, and P.~Yu, ``{BERT} post-training for review reading
  comprehension and aspect-based sentiment analysis,'' in \emph{Proceedings of
  the 2019 Conference of the North {A}merican Chapter of the Association for
  Computational Linguistics: Human Language Technologies, Volume 1 (Long and
  Short Papers)}.\hskip 1em plus 0.5em minus 0.4em\relax Minneapolis,
  Minnesota: Association for Computational Linguistics, Jun. 2019, pp.
  2324--2335. [Online]. Available:
  \url{https://www.aclweb.org/anthology/N19-1242}
\BIBentrySTDinterwordspacing

\bibitem{jeremy2018}
\BIBentryALTinterwordspacing
J.~Howard and S.~Ruder, ``Fine-tuned language models for text classification,''
  \emph{CoRR}, vol. abs/1801.06146, 2018. [Online]. Available:
  \url{http://arxiv.org/abs/1801.06146}
\BIBentrySTDinterwordspacing

\bibitem{cho2014learning}
\BIBentryALTinterwordspacing
K.~Cho, B.~van Merri{\"e}nboer, C.~Gulcehre, D.~Bahdanau, F.~Bougares,
  H.~Schwenk, and Y.~Bengio, ``Learning phrase representations using {RNN}
  encoder{--}decoder for statistical machine translation,'' in
  \emph{Proceedings of the 2014 Conference on Empirical Methods in Natural
  Language Processing ({EMNLP})}.\hskip 1em plus 0.5em minus 0.4em\relax Doha,
  Qatar: Association for Computational Linguistics, Oct. 2014, pp. 1724--1734.
  [Online]. Available: \url{https://www.aclweb.org/anthology/D14-1179}
\BIBentrySTDinterwordspacing

\bibitem{le2014distributed}
\BIBentryALTinterwordspacing
Q.~V. Le and T.~Mikolov, ``Distributed representations of sentences and
  documents,'' 2014, cite arxiv:1405.4053. [Online]. Available:
  \url{http://arxiv.org/abs/1405.4053}
\BIBentrySTDinterwordspacing

\bibitem{joachims1999making}
T.~Joachims, ``Making large-scale {SVM} learning practical,'' in \emph{Advances
  in Kernel Methods - Support Vector Learning}, B.~Sch{\"o}lkopf, C.~Burges,
  and A.~Smola, Eds.\hskip 1em plus 0.5em minus 0.4em\relax Cambridge, MA: MIT
  Press, 1999, ch.~11, pp. 169--184.

\bibitem{li2010structure}
F.~Li, C.~Han, M.~Huang, X.~Zhu, Y.~Xia, S.~Zhang, and H.~Yu, ``Structure-aware
  review mining and summarization,'' in \emph{Proceedings of the 23rd
  International Conference on Computational Linguistics (Coling 2010)}, 2010,
  pp. 653--661.

\end{thebibliography}

\end{document}